\documentclass{article}
\usepackage{spconf,amsmath,epsfig} 
\usepackage{float} 
\usepackage{enumitem}
\usepackage{booktabs}
\usepackage{multirow}
\usepackage{color}
\usepackage[numbers,compress]{natbib}

\title{LSNet: Extremely Light-weight Siamese Network for Change Detection of Remote Sensing Image}
\name{Biyuan Liu$^1$, Huaixin Chen$^{1,*}$, Zhixi Wang$^{2}$ }
\address{
	$^1$School of Resources and Environment, University of Electronic Science and Technology of China\\
	$^2$Novel Product R\&D Department, Truly Opto-Electronics Co., Ltd., Shanwei 516600, China\\
	byliu@std.uestc.edu.cn, *huaixinchen@uestc.edu.cn, wangzx.rd@trulyopto.cn}
\begin{document}
%
\maketitle
\begin{abstract}
The Siamese network is becoming the mainstream in change detection of remote sensing images (RSI). However, in recent years, the development of more complicated structure, module and training processe has resulted in the cumbersome model, which hampers their application in large-scale RSI processing. To this end, this paper proposes an extremely lightweight Siamese network (LSNet) for RSI change detection, which replaces standard convolution with depthwise separable atrous convolution, and removes redundant dense connections, retaining only valid feature flows while performing Siamese feature fusion, greatly compressing parameters and computation amount. Compared with the first-place model on the CCD dataset, the parameters and the computation amount of LSNet is greatly reduced by 90.35\% and 91.34\% respectively, with only a 1.5\% drops in accuracy.
\end{abstract}
\section{Introduction}
\label{sec:intro}

Remote sensing image (RSI) change detection is a procedure for extracting changes in the earth's surface, which is significant for map revision, agricultural investigation, and disaster assessment, etc \cite{cd_survey_rsi}. However, besides the semantic changes, which are mainly man-made changes, there are various noise changes in RSI, such as illumination, sensor noise and resolution variations \cite{ccd}. Traditional RSI change detection depends on manual features and time-consuming pre- and post-processing processes, and is difficult to distinguish semantic changes from background noise \cite{cd_survey_1}. The image pair, on the other hand, can be directly fed into the Siamese convolutional network  without preprocessing, and it relies on end-to-end supervised learning to separate semantically changed regions from invariant ones, so it has recently become the mainstream method for RSI change detection \cite{SNUNet,DASNet,cd_optical,chen2019deep,lei2019landslide,zhang2020deeply,peng2019end,FC_siamese}.

\begin{figure}[htb]
	
	\begin{minipage}[b]{1.0\linewidth}
		\centering
		\centerline{\epsfig{figure=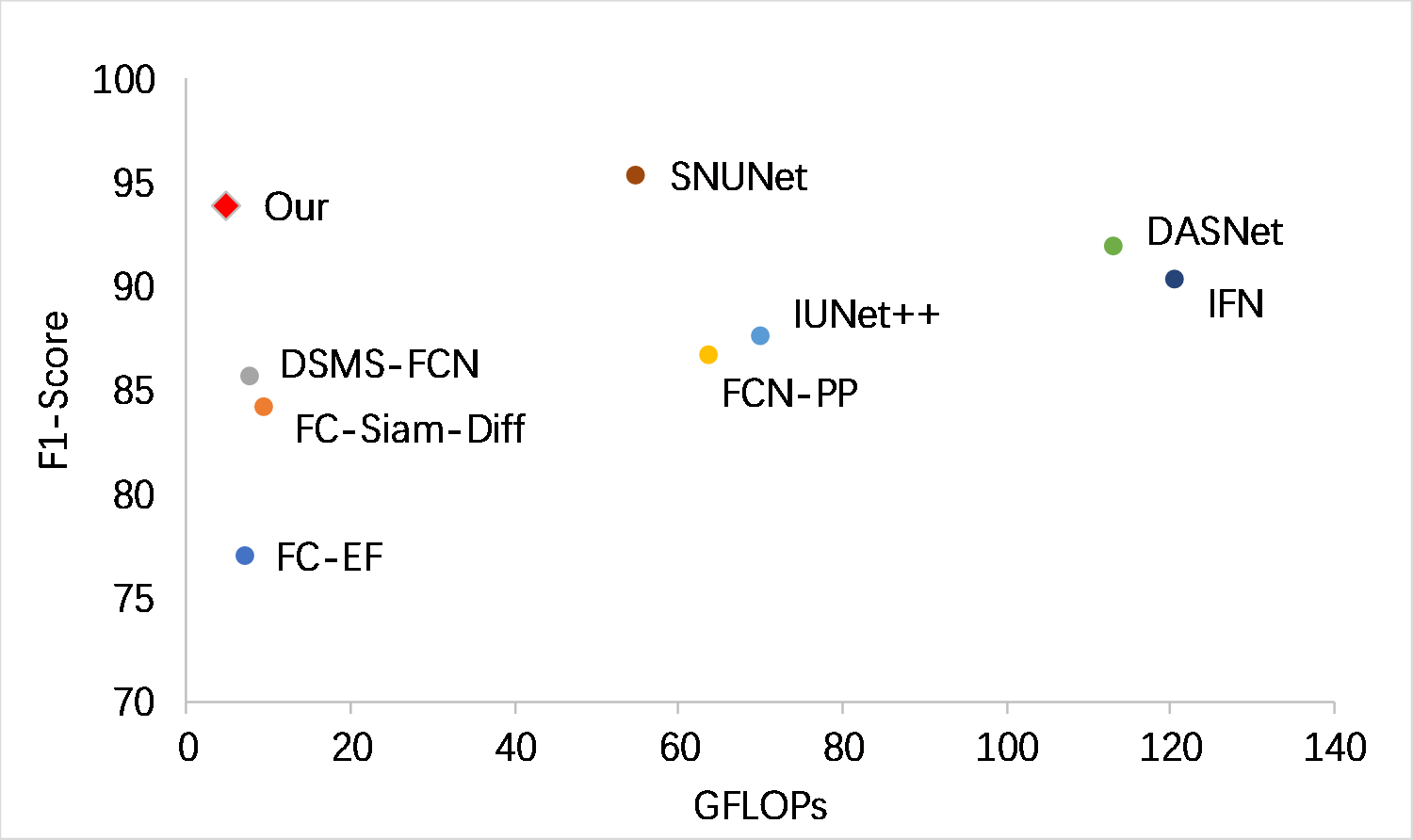,width=8.5cm}}
	\end{minipage}
	\caption{Computational (GFLOPs) $\text{vs}$ accuracy (F1-Score) of our method  and other Siamese RSI detection networks.}
	\label{fig1}
\end{figure}

\begin{figure*}[htbp]
	
	\begin{minipage}[b]{1.0\linewidth}
		\centering
		\centerline{\epsfig{figure=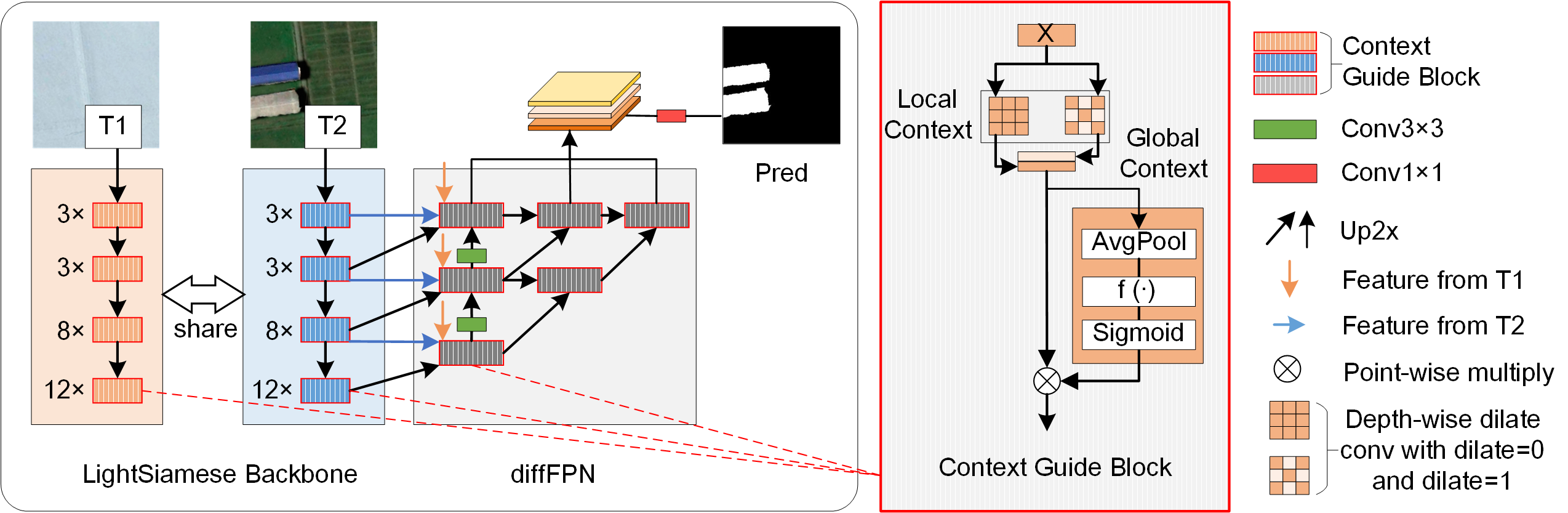,width=18cm}}
	\end{minipage}
	\caption{The pipeline of the proposed LightSiamese Network (LSNet) and the detailed structure of context guide block.}
	\label{fig2}
\end{figure*}

In order to improve the accuracy of semantic changes detection, recent research on Siamese network has primarily focused on improving the feature interaction of Siamese backbone \cite{SNUNet,lei2019landslide,peng2019end,FC_siamese}, introducing attention mechanisms \cite{SNUNet,DASNet} to refine features and improving loss function \cite{DASNet,cd_optical,chen2019deep,zhang2020deeply} to optimize the training process, but has paid little attention to the efficiency. With the rapid development of satellite observation technology (eg, QuickBird, GaoFen, and WorldView), RSI with high spatial and temporal resolution is easier to obtain. In order to handle the change detection challenge of large-scale RSI, it is necessary to improve the effectiveness of the Siamese network.

Therefore, this paper proposes an extremely lightweight Siamese network (LSNet), which outperforms existing techniques in efficiency as shown in Figure 1. The network backbone is constructed using a context-guided module composed of depthwise separable atrous convolution and global feature aggregation as the core components. Furthermore, a differential feature pyramid network (diffFPN) is presented to perform progressive feature pair difference extraction and resolution recovery, eventually separating changed from constant image areas. The main contributions can be summarized as:
\vspace{-2mm}
\begin{itemize}[leftmargin=*]
\setlength{\itemsep}{0pt}
\setlength{\parsep}{0pt}
\setlength{\parskip}{0pt}
\item	
 An extremely lightweight Siamese network for RSI change detection is constructed, which includes a 52-layers light backbone with only 3.97\% of the parameters and 32.56\% of the computation of ResNet-50 \cite{ResNet}.
\item	
The proposed diffFPN for Siamese feature fusion removes rebundant connections while maintaining valid feature flow.
\item	
Experiments on the CCD dataset show that the proposed method produces competitive results with a modest number of parameters and computation. When compared to the first-place model, the parameters and calculation amount of LSNet is only the 9.64\% and 8.67\% respectively, with only a 1.5\% decrease in accuracy.
\end{itemize}

\section{Method}
\label{sec:format}
{\vspace{-2mm}
The proposed LSNet contains a light-Siamese backbone that is built utilizing the context guide block (CGB) \cite{CGNet} intead of the commonly used residual module \cite{ResNet}, and diffFPN is used for efficient Siamese features pair fusion. Figure 2 depicts the network structure.
\vspace{-2mm}}
\subsection{Light-Siamese backbone}
The structure of basic component CGB is shown in Figure 2. The input X passes through parallel dilate convolutions \cite{deeplabv2} to obtain local context information in different ranges. The dilate convolutions here are calculated in a depthwise separable manner, i.e, all channels are grouped, and the convolution operation is only performed in an independent group \cite{Mobilenet}. Considering the input $X\in {{R}^{H\times W\times {{C}_{x}}}}$ (${\text{C}}_{\text{x}}$ is the input channels) and the output $Y\in {{R}^{H\times W\times {{C}_{y}}}}$, the calculation amount of standard convolution $K \in {R^{{{\rm{H}}_k} \times {W_k} \times {C_x} \times {C_y}}}$ used to obtain the row $h$ and column $w$ of the $cth$ channel in Y is
{\setlength\abovedisplayskip{2mm}
	\setlength\belowdisplayskip{2mm}
\begin{equation}
	Y(h,w,c)
	=\underset{i=1}{\overset{{{H}_{k}}}{\mathop{\sum }}}\,\underset{j=1}{\overset{{{W}_{k}}}{\mathop{\sum }}}\,\underset{j=1}{\overset{{{\text{C}}_{\text{x}}}}{\mathop{\sum }}}\,K(i,j,n,c)X(h+i,w+j,n)  \\
\end{equation}}
when depth-wise convolution $K\in {{R}^{{{\text{H}}_{k}}\times {{W}_{k}}\times {{C}_{x}}\times {{C}_{y}}}}$ is used, the calculation amount is
{\setlength\abovedisplayskip{2mm}
	\setlength\belowdisplayskip{2mm}
	\begin{equation}
	Y(h,w,c)
	=\underset{i=1}{\overset{{{H}_{k}}}{\mathop{\sum }}}\,\underset{j=1}{\overset{{{W}_{k}}}{\mathop{\sum }}}\,K(i,j,1,c)X(h+i,w+j,n)  \\
\end{equation}}
which is the $\text{1/}{{\text{C}}_{\text{x}}}$ of standard convolution. And then an attention module is used for channels interaction and global information extraction, which is composed of average pooling, non-linear $f(\cdot)$ and sigmoid layer and can be expressed as 
{\setlength\abovedisplayskip{2mm}
	\setlength\belowdisplayskip{2mm}
\begin{equation}
  {{Z}_{c}}=sigmoid(f(\frac{1}{H\times W}\underset{i=1}{\overset{H}{\mathop{\sum }}}\,\underset{j=1}{\overset{W}{\mathop{\sum }}}\,Y(i,j,c)))
\end{equation}}

{The two phases of images T1 and T2 pass through the light-Siamese backbone network with shared weights, which consists of 4 compound layers with $3/3/8/12$ CGBs, each CGB is equivalent to two levels, so there are 4 Group feature output, for a total of 52 layers.
\vspace{-1mm}}

\subsection{Differential feature pyramid network}
\begin{figure}[h!]
	\vspace{-2mm}
	\begin{minipage}[b]{1.0\linewidth}
		\centering
		\centerline{\epsfig{figure=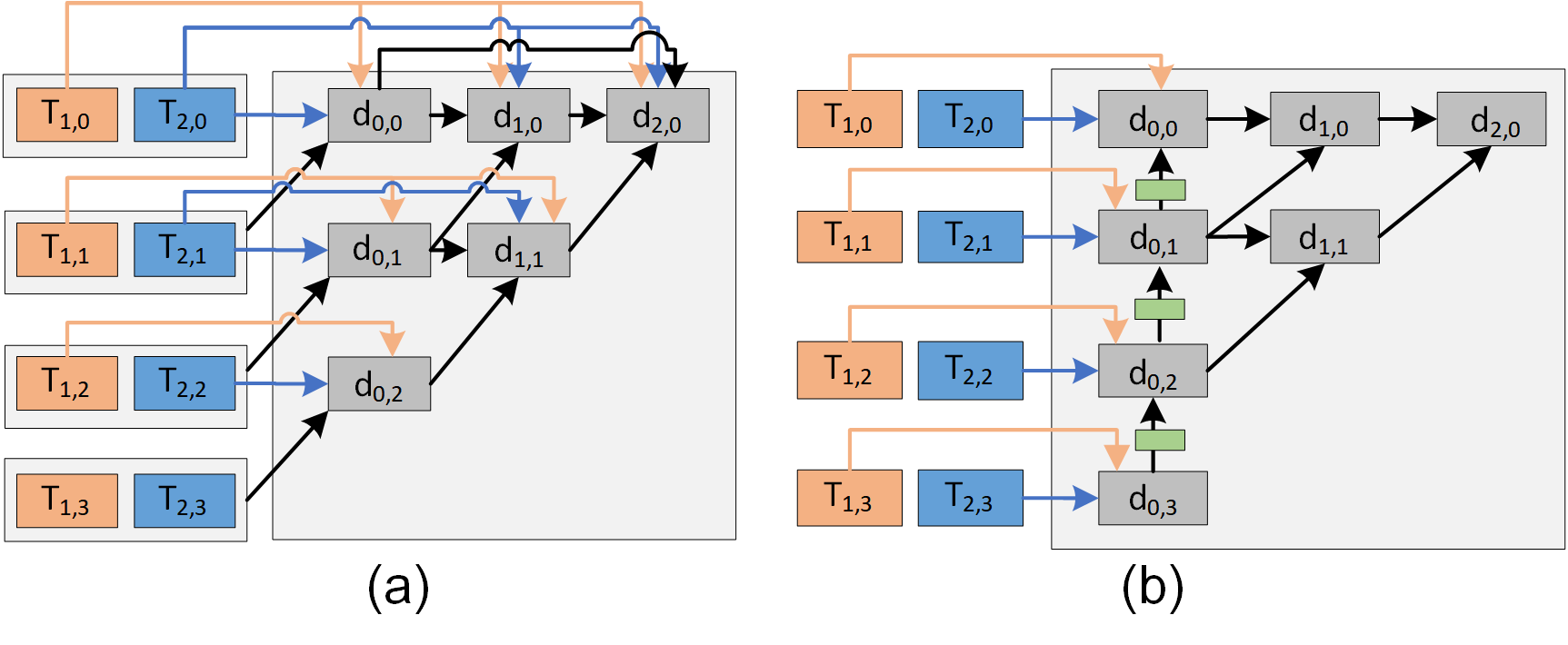,width=9.0cm}}
	\end{minipage}
	\caption{Siamese Feature Pyramid Network (FPN) Comparison. (a) denseFPN proposed in SNUNet; (b) Our diffFPN.}
	\label{fig3}
\end{figure}
\noindent A densely connected pyramid feature fusion method is proposed in SNUNet \cite{SNUNet}, as shown in Figure 3(a), and has achieved leading results. Considering three output levels ${{\text{d}}_{0,0}}\text{,}{{\text{d}}_{1,0}}\text{,}{{\text{d}}_{2,0}}$ for prediction, here the convolutional layer is abbreviated as $f(\cdot )$, then
\begin{align}
{{d}_{0,0}}&=f({{T}_{1,0}},{{T}_{2,0}},{{T}_{1,1}},{{T}_{2,1}})\\
{{d}_{1,0}}&=f({{T}_{1,0}},{{T}_{2,0}},{{d}_{0,0}},{{d}_{0,1}})\\
{{d}_{2,0}}&=f({{T}_{1,0}},{{T}_{2,0}},{{d}_{0,0}},{{d}_{1,0}},{{d}_{1,1}})
\end{align}

In this structure, there are two issues: (1) Redundant connections. Shallow features such as ${T}_{1,0},{T}_{2,0}$ are repeatedly inputted into ${\text{d}}_{0,0},{\text{d}}_{1,0}$ and ${\text{d}}_{2,0}$, which is 
inefficient. (2) Unreasonable feature flow. Despite the presence of redundant connections in denseFPN, the output layers ${{\text{d}}_{0,0}}$ and $\text{d}_{1,0}$ contains incomplete features from the backbone. The proposed diffFPN is shown in Figure 3(b), which can be stated as 
\begin{align}
	& {{d}_{0,0}}=f({{T}_{1,0}},{{T}_{2,0}},f({{d}_{0,1}})) \\ 
	& {{d}_{1,0}}=f({{d}_{0,0}},{{d}_{0,1}}) \\ 
	& {{d}_{2,0}}=f({{d}_{1,0}},{{d}_{1,1}})
\end{align}
In diffFPN, redundant connections are removed, and the added bottom-up fusion path makes the three output layers contain complete backbone network features.

\section{Experiment and Results}
\label{sec:pagestyle}
\subsection{Dataset and evaluation metrics}
RSI change detection dataset (CDD) proposed by in [9] is used for evaluation, which was obtained from Google Earth with the spatial resolution from 0.03m to 1m per pixel, and the training, validating, and testing sets contains 10000, 3000 and 3000 image-pairs with size of 256$\times$256 pixels respectively. 

For accuracy evaluation, Common accuracy including {\bf precision (P)}, {\bf recall (R)}, {\bf F1-score (F1)} and {\bf overall accuracy (OA)} are utilized. 
Previous efficiency indicators include the amount of parameters (Params) and Giga floating-point operations per second (GFLOPs). As the focus on the Siamese network effectiveness, we proposes the following normalized efficiency indicators
\vspace{0.5mm}
\begin{align}
	 &\text{F1-P}=\frac{F1}{Params/Ma{{x}_{Params}}} \\ 
	 &\text{F1-G}=\frac{F1}{GFLOPs/Ma{{x}_{GFLOPs}}} \\ 
	 &\text{F1-Eff}=(\text{F1-P}+\text{F1-G})/2 
\end{align}

$\textbf{F1-G}$ and $\textbf{F1-F}$ quantify the impact of unit parameters and calculations on the F1-score respectively, and $\textbf{F1-Eff}$, which weights $\text{F1-G}$ and $\text{F1-F}$ equally, is used to evaluate the overall efficiency of a model.

\subsection{Accuracy and efficiency comparison}

\begin{table}[tbp]
	\caption{Comparison of module parameters and computation amount. \textbf{Bold} font indicates the best results.}
	\label{table1}
	\vspace{2mm}
	\centering
	\begin{tabular}{cccc}
		\toprule
		Module & Type & Params (M) & GFLOPs \\
		\midrule
	 	\multirow{2}*{Backbone} & ResNet-50 & 23.5080 & 10.7347 \\
	 	~ & LightSiamese-52 & \bf{0.9326} & \bf{3.4956} \\
	 	\cmidrule(r){2-4}
	 	\multirow{2}*{FPN} & DenseFPN & \bf{0.1590} & 2.3348 \\
		~ & DiffFPN & 0.2299 & \bf{1.2464} \\
		\bottomrule
	\end{tabular}
\end{table}

ResNet-50 or residual block is commonly utilized in FC-EF [10], DASNet [11] and SNUNet [12], etc, whose amount of parameters and calculation is the 25 times and 3 times that of the proposed light-Siamese backbone respectively. It can be seen that the compression of parameters is more easier than it of calculation amount. As discussed in \textbf{Sec2.2}, the structure of denseFPN has irrationality of feature flow, and diffFPN is the proposed revised structure. With a small increase in the amount of parameters (0.0709M), the amount of calculation is further reduced by 1.0884 GFLOPs. 

\begin{table}[bp]
	\renewcommand\arraystretch{1.1}
	\caption{Comparison with the mainstream methods. The colors {\color{red}{red}}, {\color{green}{green}}, {\color{blue}{blue}} indicate the top 3 results for each metric.}
	\label{table1}
	\vspace{2mm}
	\centering
	\begin{tabular}{ccccc}
		\toprule
		Method&   P&   R&   OA&   F1 \\
		\midrule
		FC-EF \cite{FC_siamese}&   81.00&   73.34&   94.82&   77.00 \\
		FC-Siam-Diff \cite{FC_siamese}&   88.62&   80.29&   96.45&   84.25 \\
		DSMS-FCN \cite{chen2019deep}&   89.34&   82.40&   96.85&   85.73 \\
		FCN-PP \cite{lei2019landslide}&   91.77&   82.21&   97.03&   86.73 \\
		IUNet++ \cite{peng2019end}&   89.54&   87.11&   96.73&   87.56 \\
		DASNet \cite{DASNet}&   92.52&   91.45&   98.07&   91.93 \\
		IFN \cite{zhang2020deeply}&   \color{green}{94.96}&   86.08&   97.71&   90.30 \\
		SNUNet \cite{SNUNet}&   \color{red}{95.6}&   \color{red}{94.9}&   -&   \color{red}{95.3} \\
		Our-denseFPN&   93.84&   \color{green}{94.19}&   \color{blue}{98.52}&   \color{blue}{94.02} \\
		Our-diffFPN&   \color{blue}{94.31}&   \color{blue}{93.95}&   \color{green}{98.55}&   \color{green}{94.13} \\
		\bottomrule
	\end{tabular}
\end{table}

In Table 2, the results of the proposed LSNet outperforms alternative networks using ResNet-50, indicating the high efficiency of the proposed backbone. The structural improvement from denseFPN to diffFPN improves the results of many indicators such as P, OA, and F1. And majority of the proposed LSNet'indicators  are in the second and third places.

\begin{figure*}[htbp]
	\centering
	\begin{minipage}[t]{0.48\textwidth}
		\centering
		\includegraphics[width=8.5cm]{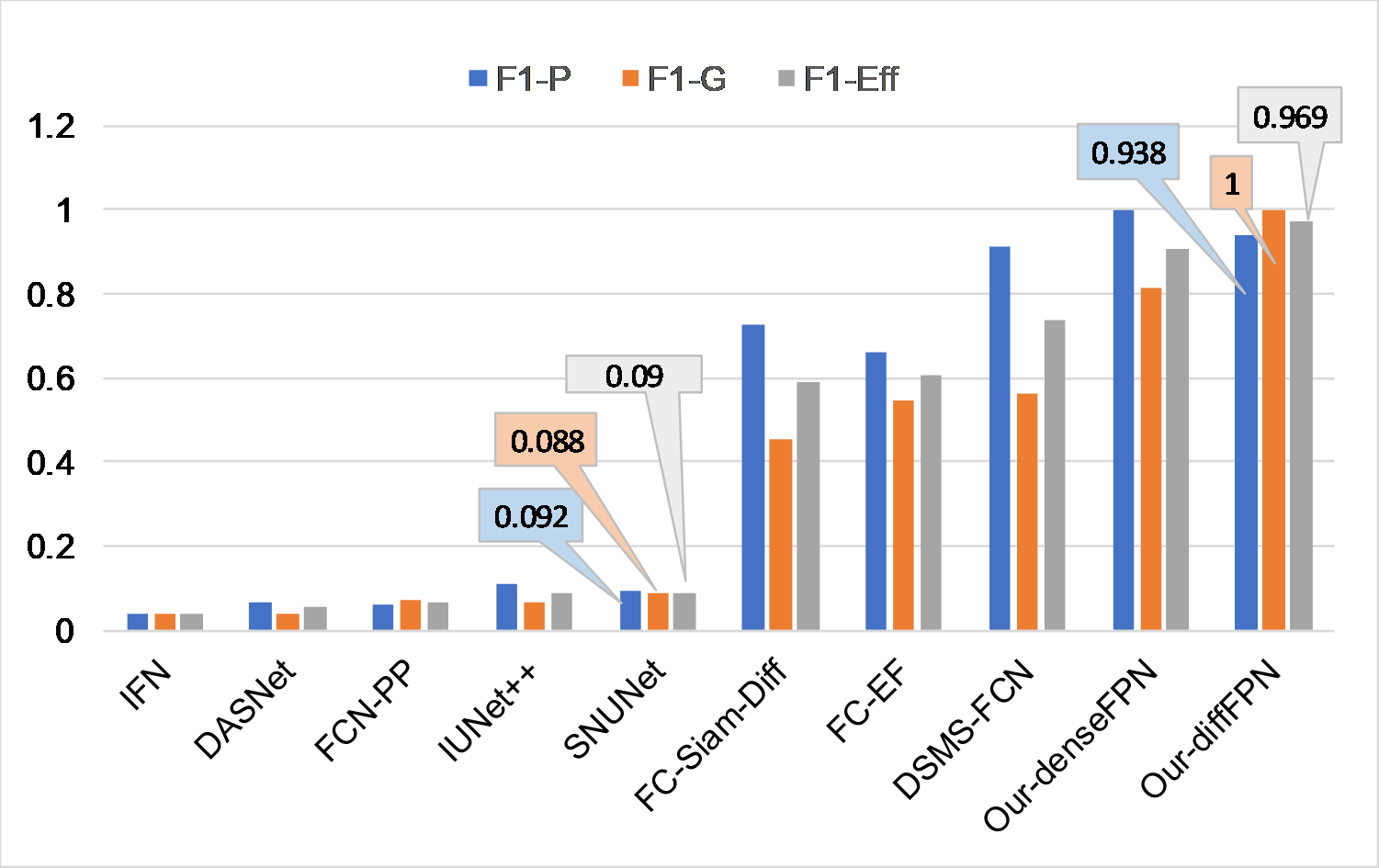}
		\caption{Efficiency comparision with F1-P, F1-G and F1-Eff}
	\end{minipage}
	\hspace{5mm}
	\begin{minipage}[t]{0.48\textwidth}
		\centering
		\includegraphics[width=8.5cm]{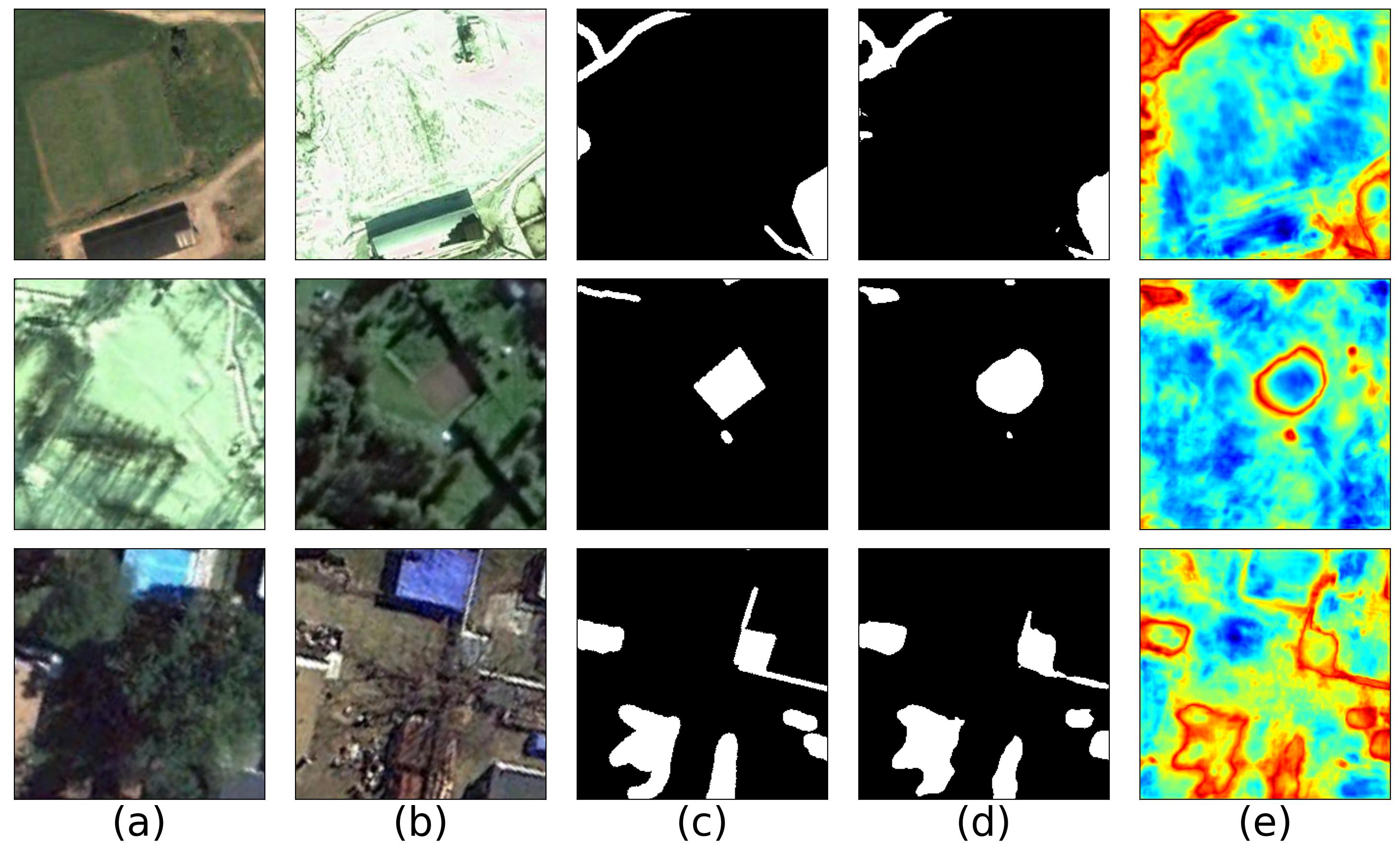}
		\caption{Visualization results. (a) Image in T1. (b) Image in T2. (c) Ground truth. (d) Prediction. (e) Class score map.}
	\end{minipage}
\end{figure*}

Figure 4 further compares the the efficiency. It can be seen that our method with diffFPN has the highest F1-Eff and F1-G. Compared to SNUNet, the method with highest accuracy, the amount of parameters and calculation of our LSNet has reduced by 90.35\% and 91.34\% respectively, with just a 1.5\% decline in accuracy.

Figure 5 shows that the ground cover has changed greatly between T1 and T2. Even the constant buildings, their colors and textures have also changed, retaining only the basic structures. Changes defined in this dataset are mainly caused by human activity. Comparing column (c) with column (d), the detection results of LSNet are relatively accurate. From column (e), it can be found that the edge of the changing area has higher probability than the inner areas, which indicates that the network leverages the structure of the area as a discriminant feature, which improves its robustness to color and texture changes.

\section{Conclusion}
For effictive RSI change detection, A lightweight Siamese network with a context-guided block based backbone and feature pair fusion module is proposed. The results on the challenging CCD dataset reveal that when compared to other mainstreams, the proposed method achieves competitive results with a limited amount of parameters and computations, demonstrating its efficiency.

\textbf{Acknowledgements} This paper is supported by the "YangFan" major project in Guangdong province of China, No. [2020] 05.

\bibliographystyle{IEEEbib}
\bibliography{strings,refs}

\end{document}